\documentclass{article} 
\usepackage{iclr2027_conference,times}

\usepackage{amsmath,amsfonts,bm}

\def\eqref#1{equation~\ref{#1}}

\def\1{\bm{1}}

\DeclareMathAlphabet{\mathsfit}{\encodingdefault}{\sfdefault}{m}{sl}
\SetMathAlphabet{\mathsfit}{bold}{\encodingdefault}{\sfdefault}{bx}{n}

\usepackage{hyperref}
\usepackage{url}
\usepackage{amssymb}
\usepackage{booktabs}
\usepackage{multirow}
\usepackage{makecell}
\usepackage[most]{tcolorbox}
\usepackage{listings}

\newcommand{\meanstd}[2]{#1 {\scriptsize\normalfont$\pm$ #2}}

\title{Constrained Decoding: Harmful or Beneficial? Revisiting Its Impact on LLM Performance}

\author{
\textbf{Ngoc Trinh Hung Nguyen}\textsuperscript{1,2},
\textbf{Alonso Silva}\textsuperscript{2},
\textbf{Armen Aghasaryan}\textsuperscript{2},
\textbf{Mehwish Alam}\textsuperscript{1}
\\
\textsuperscript{1}Télécom Paris, Institut Polytechnique de Paris, France
\\
\textsuperscript{2}Nokia Bell Labs
\\
\\
\small{
\textbf{Correspondence:}
\href{mailto:ngoc.nguyen@ip-paris.fr}{ngoc.nguyen@ip-paris.fr}
}
}

\iclrfinalcopy 
\begin{document}

\maketitle

\begin{abstract}
Constrained decoding guarantees that Large Language Models (LLMs) produce outputs conforming to predefined formats, but is often assumed to inherently degrade reasoning. We examine this assumption through a taxonomy of three constraint types: \emph{format constraints}, which enforce output structure; \emph{validity constraints}, which mask invalid answer; and \emph{reasoning constraints}, which restrict reasoning trajectories. We theoretically show that appropriate constraints can preserve or improve performance by eliminating invalid paths. Across four model families and eleven open-source models, causal ablations on diverse benchmarks using state-of-the-art constrained decoding strategies reveal that apparent performance degradation largely stems from non-equivalent prompting and output formats rather than from decoder-level masking. Meanwhile, validity constraints improve reliability and extractability by eliminating invalid answers, while reasoning constraints can improve reasoning and, consequently, accuracy when they enforce beneficial trajectories. These findings show that the impact of constrained decoding depends on the type of constraint and the output space it removes. Our code is available online\footnote{The code is included in the Supplementary Materials and will be publicly released upon acceptance.}.

\end{abstract}

\section{Introduction}

LLMs have shown strong performance in tasks such as text completion, question answering, code generation, and data extraction \citep{minaee2024large}. Their capabilities can be extended by incorporating external systems through function calling, tool use, and agentic frameworks using JSON schemas or other formal formats \citep{schick2023toolformer, patil2024gorilla}. However, LLMs do not always produce structurally valid outputs, creating a growing need for reliable format control in industrial applications \citep{liu2024we}. To address this gap, grammar-constrained decoding methods (\citep{willard2023efficient, koo2024automata, dong2024xgrammar,guidance2025}) mask invalid tokens during generation, ensuring syntactically valid outputs. 

Recent observations show that such strict constraints can negatively affect LLM reasoning, leading to a significant drop in accuracy \citep{tam-etal-2024-speak, banerjee2025crane, nguyen2026thinking, reddy2026hidden, lee2026format}. Several studies have investigated the causes of this degradation. Most studies attribute the performance drop to constrained decoding and propose solutions such as adding a \texttt{reason} field to JSON schema, two-step generation-then-formatting approaches \citep{tam-etal-2024-speak, reddy2026hidden}, and free reasoning followed by constrained decoding for final formatting via trigger tokens \citep{nguyen2026thinking}. Other studies, however, suggest that the drop is mainly caused at prompt-level \citep{lee2026format} or unfair comparisons between constrained and unconstrained settings \citep{dottxt2024say}.

To investigate these different perspectives, we separate constraints into three types: \emph{format constraints}, which enforce a specific output format, such as JSON; \emph{validity constraints}, which mask invalid continuations based on requirements such as data types (e.g., integers or strings), length; and \emph{reasoning constraints}, which restrict the model's reasoning process.

Theoretically, we show that constrained decoding does not harm LLM generation when the prompt and expected output format\footnote{The term \emph{expected output format} refers to the format explicitly required by the prompt and accepted by the evaluation; see Section~\ref{sec:prompt-format-constraints} for details.} remain unchanged. We further show that \emph{format constraints} (FC) guarantee the expected output format; \emph{validity constraints} (VC) provide a near-guaranteed, near-zero-cost gain by eliminating invalid outputs based on the task structure; and \emph{reasoning constraints} (RC) can improve performance when the enforced reasoning trajectory is beneficial.


Experiments on eleven open-source models ranging from 1.7B to 32B across four model families---Qwen \citep{qwen3.5}, Gemma \citep{team2024gemma}, Llama \citep{dubey2024llama}, and SmolLM \citep{bakouch2025smollm3}---suggest that, while a noticeable performance gap exists across different output formats, as observed in prior work \citep{tam-etal-2024-speak, banerjee2025crane, nguyen2026thinking, reddy2026hidden, lee2026format}, this gap is mainly caused by prompt-level differences rather than constrained decoding \citep{lee2026format}. Under an apples-to-apples comparison with the same prompt and expected output format, constrained decoding consistently matches or outperforms unconstrained decoding, with validity constraints providing the most consistent gains (up to 17.4\%) and reasoning constraints providing gains only when the enforced trajectory is itself beneficial.

Our main contributions are: (1) a taxonomy of three types of constraints: \emph{format constraints} (FC), \emph{validity constraints} (VC), and \emph{reasoning constraints} (RC), which characterizes how constrained decoding can benefit LLM generation beyond format guarantees; (2) a theoretical proof that, under the same prompt and expected output format, constrained decoding does not harm LLM generation; (3) a comprehensive evaluation across models from different families and sizes and benchmarks ranging from standard reasoning tasks to competition-level problems, demonstrating the benefits of constrained decoding; (4) an extended dataset that further evaluates these benefits and shows that they remain valuable even for models of substantial size; and (5) a comprehensive analysis of the sources of performance gains from constrained decoding.

This paper is structured as follows: Section~\ref{sec:related_work} reviews related work, Section~\ref{sec:methodology} presents a theoretical proof of why constrained decoding is beneficial under the same prompt and expected output format, and Sections~\ref{sec:experiments} and \ref{sec:results_discussion} present the experimental setup, results and discussion, respectively.

\section{Related Work}\label{sec:related_work}

Constrained decoding \citep{willard2023efficient,koo2024automata,dong2024xgrammar,guidance2025} guarantees the required format without the need for fine-tuning by controlling the generation process from start to end. Constrained decoding is often assumed to be limited to structured formats such as JSON, driven by its common practical use cases.

\citet{tam-etal-2024-speak} show the performance degradation of constrained decoding compared with unconstrained decoding, specifically when reasoning within a JSON schema rather than reasoning freely. \citet{banerjee2025crane} prove that finite output grammars without reasoning reduce model expressivity, while \citet{reddy2026hidden} attribute the degradation to distribution shift caused by constrained generation that enforces the output format. 

Several strategies have been proposed to allow LLMs to reason while still producing an extractable format, ranging from adding a \texttt{reason} field to JSON, to two-step free reasoning followed by formatting, \textbf{NL-to-Format} \citep{tam-etal-2024-speak,reddy2026hidden}, or single-pass free reasoning followed by formatting through a trigger token mode switch, \textbf{InWriting} \citep{nguyen2026thinking}

Our theoretical analysis shows that, when the prompt and expected output format are aligned with the grammar, constrained decoding excludes only invalid generations and should therefore not impair model performance. Empirically, while reasoning within JSON can underperform free-form reasoning, we find that, under matched prompts and output formats, constrained decoding consistently outperforms unconstrained decoding across the state-of-the-art (SOTA) decoding strategies we evaluate.

\section{Methodology}\label{sec:methodology}

In this section, we first study when constrained decoding can help LLMs produce better answers. In particular, we show that constrained decoding can improve answers both syntactically and semantically by removing invalid outputs and shifting probability toward valid ones.

We then study whether constrained decoding can also improve reasoning. We show that this is possible, but requires a suitable prior condition on the reasoning traces allowed by the constraint.

\subsection{Constrained Decoding Can Help LLMs Produce Better Answers}\label{sec:constrained-decoding-answering}

We first show that, under the same conditions\footnote{The same prompt, expected output format, and reasoning traces}, applying constrained decoding to the answer generally leads to better performance than unconstrained decoding.

Let $x$ be the input prompt, and $R$ be a fixed reasoning trace.

Let $\mathcal{Y}$ be the set of all possible candidate final-answer token
sequences of length $T$, generated from vocabulary $V$.

Let $Y^* \in \mathcal{Y}$ be the ground-truth correct final answer

Let $\mathcal{Y}_C \subseteq \mathcal{Y}$ denote the subset of all candidate answers that satisfy the predefined constraint $C$.

Let $Y$ be a token sequence $s = (s_1,\ldots,s_T)$ representing the final answer, generated given the input prompt $x$ and reasoning trace $R$. Since the model is autoregressive,
\begin{equation}
\label{eq:eq1}
P_\theta(Y \mid x, R)
= \prod_{t=1}^{T} P_\theta(s_t \mid x, R, s_{<t}).
\end{equation}

At each step, constrained decoding masks invalid next tokens given the
current prefix and renormalizes the distribution locally:

\begin{equation}
P_{\mathcal{C}}(s_t \mid x, R, s_{<t})
=
\frac{
P_\theta(s_t \mid x, R, s_{<t})
\mathbf{1}\!\left[s_t \in V_C(s_{<t})\right]
}{
Z_t(x, R, s_{<t})
},
\end{equation}

where
\begin{equation}
Z_t(x, R,s_{<t}) = \sum_{s'_t \in V_C(s_{<t})} P_\theta(s'_t \mid x, R,s_{<t})
\label{eq:mass}
\end{equation}

and $V_C(s_{<t})$ denotes the set of tokens allowed by $C$ at that position.

At each step, constrained decoding masks invalid tokens and renormalizes the remaining distribution. We assume that, for every $t$, $V_C(s_{<t})$ contains at least one token with positive probability. Otherwise, $Z_t(x, R, s_{<t}) = 0$, meaning that no permissible token can be sampled at step $t$. Thus,
\begin{equation}
0 < Z_t(x, R, s_{<t}) \leq 1
\qquad
\text{for all } t, x, R, s_{<t}.
\label{eq:normalization_bound}
\end{equation}

Therefore, the probability of the complete constrained answer $Y$ can
be written as

\begin{equation}
P_{\mathcal{C}}(Y \mid x, R) = \prod_{t=1}^{T} P_{\mathcal{C}}(s_t \mid x, R, s_{<t}) = \frac{P_\theta(Y \mid x, R)\mathbf{1}\!\left[Y \in \mathcal{Y}_C\right]}{\displaystyle\prod_{t=1}^{T} Z_t(x, R, s_{<t})}.
\label{eq:constrained_sequence_probability}
\end{equation}

as the product of indicators is equivalent to a single
sequence-level constraint:
\begin{equation}
\prod_{t=1}^{T}
\mathbf{1}\!\left[s_t \in V_C(s_{<t})\right]
=
\mathbf{1}\!\left[Y \in \mathcal{Y}_C\right].
\end{equation}

\textbf{Accuracy Metric:}

Define the accuracy indicator function $g: \mathcal{Y} \to \{0, 1\}$ as:$$g(Y) = \mathbb{I}(Y = Y^*) = \begin{cases} 1 & \text{if } Y = Y^* \\ 0 & \text{if } Y \neq Y^* \end{cases}$$

The expected accuracy without constrained decoding is:
\begin{equation}
\mathbb{E}_{Y \sim P}[g(Y) \mid R,x]
=
\sum_{Y \in \mathcal{Y}} P_\theta(Y \mid R,x)\,\mathbb{I}(Y=Y^*)
=
P_\theta(Y^* \mid R,x).
\label{eq:unconstrained_expected_accuracy}
\end{equation}

The expected accuracy with constrained is:
\begin{equation}
\mathbb{E}_{Y \sim P_{\mathcal{C}}}[g(Y) \mid R, x]
=
\sum_{Y \in \mathcal{Y}_C}
P_{\mathcal{C}}(Y \mid R, x) \cdot \mathbb{I}(Y = Y^*)
\label{eq:constrained_expected_accuracy}
\end{equation}

\subsubsection{When constrained decoding does not hurt LLMs Performance}

\textbf{Assumption 1 -- Soundness of output constraints.}
The ground-truth answer $Y^*$ satisfies\footnote{Here, ``satisfies'' means that $Y^*$ is both semantically correct and syntactically valid under $C$. Thus, a semantically correct answer that cannot be extracted according to $C$ is considered incorrect, details in Sec.~\ref{sec:sample-level-failed-parse}.} the output constraint $C$.

\begin{equation}
Y^* \in \mathcal{Y}_C .
\label{eq:assumption_soundness}
\end{equation}

From Eq.~\ref{eq:normalization_bound}, we have
\begin{equation}
0 < \prod_{t=1}^{T} Z_t(x,R,s_{<t}) \leq 1.
\qquad
\text{for all } t, x, R, s_{<t}.
\label{eq:product_normalization_bound}
\end{equation}
Therefore, under Assumption~1, Eq.~\ref{eq:constrained_expected_accuracy} becomes
\begin{equation}
\mathbb{E}_{Y \sim P_{\mathcal{C}}}[g(Y) \mid R,x]
=
\frac{P_\theta(Y^* \mid R,x)}
{\displaystyle\prod_{t=1}^{T} Z_t(x,R,s^*_{<t})}
\geq P_\theta(Y^* \mid R,x).
\label{eq:constrained_accuracy_bound}
\end{equation}

where $s^*_{<t} = (s^*_1, \ldots, s^*_{t-1})$ denotes the prefix of length $t-1$ belonging to the ground-truth sequence $Y^* = (s^*_1, \ldots, s^*_T)$.

Assumption~1---the condition that $C$ does not hurt LLM performance---is usually satisfied in practice by defining $C$ to enforce only the output format while allowing the answer field to take any value. It can fail when $C$ semantically excludes the correct ground-truth answer.

\subsubsection{When constrained decoding improve LLMs Performance}

For the ground-truth answer $Y^*=(s^*_1,\ldots,s^*_T)$, define the \emph{path factors}
\begin{equation}
Z^*_t := Z_t(x,R,s^*_{<t}),
\qquad
Z^* := \prod_{t=1}^{T} Z^*_t .
\label{eq:path_factor}
\end{equation}
Thus, $Z^*_t$ is the total unconstrained probability mass of the tokens
allowed by $C$ after the prefix $s^*_{<t}$, and $Z^*$ is the product of these
masses along the path of $Y^*$.

\textbf{Assumption 2 -- Non-trivial masking on the path of $Y^*$.}

There exists a step $t$ and a token $v \neq s^*_t$ such that
\[
v \notin V_C(s^*_{<t})
\quad\text{and}\quad
P_\theta(v \mid x,R,s^*_{<t}) > 0 .
\]
That is, at some prefix of $Y^*$, the mask removes an invalid token
to which the model assigns positive probability.

Under Assumption~2,
\begin{equation}
Z^*_t
=
1-\sum_{v\notin V_C(s^*_{<t})}
P_\theta(v\mid x,R,s^*_{<t})
<1,
\qquad \text{for that step } t.
\label{eq:strict_path_factor}
\end{equation}

Since every other factor is at most $1$, $Z^* < 1$. Therefore, under Assumptions~1 and~2,
\begin{equation}
\mathbb{E}_{Y \sim P_{\mathcal{C}}}[g(Y)\mid R,x]
= \frac{P_\theta(Y^*\mid x,R)}{Z^*} > P_\theta(Y^*\mid x,R).
\label{eq:strict_accuracy_bound}
\end{equation}

Conversely, if the constraint masks no token with positive probability along the path of $Y^*$, then $Z^*=1$ and the two expectations are equal.

Equation~\ref{eq:strict_path_factor} clarifies when constrained decoding
can provide a benefit:
\begin{enumerate}
    \item \textbf{Format constraint.} If the model already follows $C$, the constraint has no effect. If the model assigns positive probability to syntactically invalid tokens, such as a missing ``\texttt{\{}'' or ``\texttt{"}'' required by $C$, constrained decoding removes this probability mass and increases the relative probability of $Y^*$.

    \item \textbf{Validity constraint.} If $C$ additionally incorporates
    semantic information about the ground-truth answer, more invalid tokens
    can be ruled out along the path of $Y^*$, making $Z^*$ smaller and the
    resulting gain larger.
\end{enumerate}


\subsection{Can Constrained Decoding helps LLMs become a Better Reasoner}

We extend the analysis of constrained decoding to reasoning traces.

Let $Y^*$ denote the ground-truth answer, and $R^*$ denote the set of reasoning traces that lead to $Y^*$. 

Let $C$ denote the constraint set over reasoning--answer pairs $(R,Y)$. We consider three cases:

\textbf{Case 1 --- Total exclusion (hard elimination).} If $C$ completely excludes the good reasoning traces and correct answer,
\[
(R^*, Y^*) \cap C = \varnothing,
\]

Then the constraint admits neither better reasoning traces $R^*$ nor the correct answer $Y^*$. Hence, no correct reasoning--answer pair can be generated, yielding zero expected correctness. Thus, constrained decoding strictly hurts performance whenever the unconstrained model assigns positive probability to a correct pair.

\textbf{Case 2 --- Oracle constraint (perfect retention).} If $C$ retains only the good reasoning traces and correct answer,
\[
C = \{(R^*,Y^*)\},
\]

then constrained decoding always generates a correct reasoning--answer pair. Therefore, it achieves perfect correctness and strictly outperforms unconstrained decoding whenever the latter assigns positive probability to an incorrect pair.

\textbf{Case 3 --- Partial retention (practical case).}

The constraint retains some good reasoning--answer pairs while excluding others,
which may include both good and bad pairs. Formally,
\[
\begin{aligned}
(R^*,Y^*)\cap C &\neq \varnothing, &
(R^*,Y^*) &\not\subseteq C,\\
C\setminus(R^*,Y^*) &\neq \varnothing. &&
\end{aligned}
\]

Thus, both favorable and unfavorable reasoning--answer pairs remain possible, and the effect on expected correctness is indeterminate in general. 

As shown in Equation~\ref{eq:strict_path_factor}, constrained decoding is more effective when we have greater knowledge of the reasoning paths that lead to the correct answer. Its goal is therefore to guide the model toward better reasoning--answer pairs, although such improvement is not guaranteed because the quality of the resulting reasoning cannot generally be predicted a priori.

\section{Experimental Setup}\label{sec:experiments}
\subsection{Models and Datasets}
\textbf{Models}

We evaluate eleven open-source models in total. Ten models: Qwen3-1.7B, Qwen3-4B, Qwen3-8B, Qwen3.5-2B, Qwen3.5-4B, Qwen3.5-9B, Qwen3.5-27B, Llama3-8B, Gemma2-9B, and SmolLM3-3B, are evaluated for task solving, while Qwen3-32B is used as the extractor model in the two-step NL-to-Format approach. All models were released between 2024 and 2026. Detailed configurations are provided in Sec.~\ref{sec:experimental-septup}.

\textbf{Datasets}

The datasets span two levels of reasoning difficulty. Standard benchmarks include \textbf{GSM8K} \citep{cobbe2021training}, \textbf{Last Letter Concatenation} \citep{wei2022chain}, which cover multi-step mathematical and symbolic reasoning tasks for which prior work \citep{tam-etal-2024-speak} has reported performance degradation under constrained decoding.


Reasoning-intensive benchmarks include \textbf{AIME 2026}, challenging competition-level mathematics problems \citep{dekoninck2026matharena}, and \textbf{LastLetter-Extended}, our extension of Last Letter Concatenation from four to eight names. Both were released or constructed after the release of all models in our experiments, reducing the potential for data leakage. Dataset construction details are provided in Section~\ref{subsubsec:dataset-construction}.

\subsection{Evaluation}

We report \textbf{accuracy} as the primary metric, along with four additional metrics: \textbf{parse rate}, \textbf{format gain}, \textbf{validity gain}, and \textbf{reasoning gain}.

\paragraph{Accuracy Metrics.}
Accuracy is computed by exact match between extracted and ground-truth answers. We report mean $\pm$ std across seeds and task/format-prompt combinations (see Sec.~\ref{sec:prompt-format-constraints})

\paragraph{Parse Metrics.}
Parse rate measures the proportion of generated outputs that can be successfully parsed according to the specified format schema.

\paragraph{Format Gain.}
Format gain measures the improvement obtained by enforcing the required output format through constrained decoding. We denote this setting as \textbf{+FC}, indicating that \emph{format constraints} (\textbf{FC}) are applied.

\paragraph{Validity Gain.}
Validity gain measures the improvement obtained by incorporating additional information about the viable answer subspace into the predefined valid schema. We denote this setting as \textbf{+VC}, indicating that \emph{validity constraints} (\textbf{VC}) are applied.

\paragraph{Reasoning Gain.}
Reasoning gain measures the improvement in accuracy obtained by enforcing beneficial trajectories through constrained decoding. We denote this setting as \textbf{+RC}, indicating that \emph{reasoning constraints} (\textbf{RC}) are applied.

\subsection{Output Formats, Prompts Variations and Decoding Strategies}

We experiment with current SOTA decoding strategies that support reasoning with constrained decoding: (1) \textbf{JSON}, which adds a \texttt{reason} field to the JSON output; (2) \textbf{NL-to-Format}, which uses two-step free reasoning followed by formatting \citep{tam-etal-2024-speak,reddy2026hidden}; and (3) \textbf{InWriting}, which uses single-pass free reasoning followed by formatting through a trigger-token mode switch \citep{nguyen2026thinking}. Each strategy specifies an expected output format in the prompt and a corresponding schema for constrained decoding. Details are provided in Sec.~\ref{sec:prompt-format-constraints}.

We apply constrained decoding using Litelines, following prior work \citep{nguyen2026thinking}. Litelines allows models to perform free-form reasoning before switching to constrained decoding through trigger tokens and is built on \texttt{outlines-core}, the same underlying library used by Outlines \citep{willard2023efficient}. The format schemas and Litelines configurations are detailed in Sec.~\ref{sec:experimental-septup}.

In reporting our experimental results, we use the decoding method name to denote the expected output format. For unconstrained decoding, we append \textbf{Unconstr.}. For constrained decoding, we append the types of constraints applied, using \textbf{+FC}, \textbf{+VC}, and \textbf{+RC} for format, validity, and reasoning constraints, respectively. Multiple constraints combine labels, such as \textbf{+FC+VC}. 

\section{Results and Discussion}\label{sec:results_discussion}

In this section, we first show that (1) constrained decoding is not limited to the JSON format, (2) under the same prompt and expected output format, constrained decoding consistently outperforms unconstrained decoding, not only by enforcing the output format but also by validating answers, and (3) constrained decoding can improve LLM performance by enforcing beneficial reasoning traces. 

We then provide additional analyses of (4) the parsability of different decoding strategies, including JSON, InWriting, and NL-to-Format, (5) the sources of performance gains from Format Constraints (FC), Validity Constraints (VC), and Reasoning Constraints (RC), and (6) token efficiency across different decoding strategies.

\subsection{Constrained Decoding Beyond JSON}

Constrained decoding is not limited to JSON or other strict formats; it is a general logit-masking mechanism that also applies to plain-text generation. To demonstrate this, we compare it with the two-step NL-to-Format approach of \citet{tam-etal-2024-speak}, which uses a second LLM to extract answers from the first LLM's reasoning as plain text. We adopt the same setup but apply constrained decoding to the second-step plain-text generation. As shown in Table~\ref{tab:NL-to-Format-direct-comparison} and \ref{tab:NL-to-Format-direct-comparison-full}), constrained decoding consistently outperforms unconstrained decoding, even for plain-text generation.

Few-shot examples improve pattern adherence and string-based extraction (see \textbf{NL (Baseline)} in Table~\ref{tab:NL-to-Format-direct-comparison-full}), but offer limited reasoning benefits, consistent with prior work \citep{tam-etal-2024-speak,nguyen2026thinking}. They are also impractical for reasoning-intensive tasks such as AIME2026. We therefore use zero-shot settings for all subsequent experiments.

\begin{table}[t]
\caption{LastLetter Task: comparison of unconstrained and combined Format and Validity Constraints (FC+VC) for plain-text answers using the NL-to-Format decoding strategy with the same prompt and reasoning traces. \textit{Italicized} results are reported by \citet{nguyen2026thinking}, while \textbf{Bold} values indicate the highest accuracy.}
\label{tab:NL-to-Format-direct-comparison}
\centering
\small
\begin{tabular}{lll|ll}
\textbf{Model} &
\textbf{NL (Baseline)} &
\makecell{\textbf{JSON}\\\textbf{+FC+VC}} &
\makecell{\textbf{NL-to-Format}\\\textbf{Unconstr. (plain-text)}} &
\textbf{+FC+VC} \\

\midrule

 Qwen3-1.7B   & \textit{\meanstd{33.6}{21.9}} & \textit{\meanstd{45.4}{19.1}} & \textit{\meanstd{60.7}{4.3}} & \textbf{\meanstd{66.1}{4.8}} \\
 Qwen3-4B     & \textit{\meanstd{32.1}{26.2}} & \textit{\meanstd{41.9}{21.6}} & \textit{\meanstd{57.5}{7.5}} & \textbf{\meanstd{60.4}{7.1}} \\
 Qwen3-8B     & \textit{\meanstd{63.8}{17.4}} & \textbf{\textit{\meanstd{78.4}{7.9}}}  & \textit{\meanstd{69.4}{11.4}} & \meanstd{72.4}{10.6} \\
 Qwen3.5-2B   & \textit{\meanstd{22.4}{29.9}} & \textbf{\textit{\meanstd{69.1}{2.2}}}  & \textit{\meanstd{62.9}{5.6}} & \meanstd{68.7}{5.2}\\
 Qwen3.5-4B   & \textit{\meanstd{81.6}{1.6}}  & \textit{\meanstd{71.6}{28}}   & \textit{\meanstd{84.1}{2.3}} & \textbf{\meanstd{86.9}{2.6}}\\
 Qwen3.5-9B   & \textit{\meanstd{85.9}{1.2}}  & \textit{\meanstd{77}{27}}     & \textit{\meanstd{86.8}{1.2}} & \textbf{\meanstd{89.9}{0.5}} \\

\end{tabular}
\end{table}

\subsection{JSON as a Practical Interface for Information Extraction}

While constrained decoding is not specific to JSON, JSON provides a structured interface for extracting and processing multiple fields. We thus compare NL-to-Format with plain text (original) and JSON. As shown in Table~\ref{tab:plain-text-vs-json}, JSON performs comparably to or better than plain text across all models, while using the same free-reasoning step (Natural Language—NL). Since structured outputs become more useful as the number of fields increases, we use JSON as the target format for the remaining experiments.
\begin{table}[t]
\caption{LastLetter Task: comparison of plain-text and JSON answer formats using the NL-to-Format decoding strategy with the same free-reasoning (NL) traces, under unconstrained decoding and combined Format and Validity Constraints (FC+VC). \textbf{Bold} values indicate the highest accuracy.}
\label{tab:plain-text-vs-json}
\centering
\small
\begin{tabular}{l|ll|ll}
\textbf{Model} &
\makecell{\textbf{NL-to-Format}\\\textbf{Unconstr. (plain-text)}} &
\textbf{+FC+VC} &
\makecell{\textbf{NL-to-Format}\\\textbf{Unconstr. (JSON)}} &
\textbf{+FC+VC} \\
\midrule

Qwen3-1.7B  & \meanstd{60.7}{4.3}  & \meanstd{66.1}{4.8}  & \meanstd{67.3}{4.1}  & \textbf{\meanstd{68.9}{4.6}} \\
Qwen3-4B    & \meanstd{57.5}{7.5}  & \meanstd{60.4}{7.1}  & \meanstd{60.3}{6.4}  & \textbf{\meanstd{61.4}{6.5}} \\
Qwen3-8B    & \meanstd{69.4}{11.4} & \meanstd{72.4}{10.6} & \meanstd{72.1}{10.5} & \textbf{\meanstd{73.5}{10.6}} \\
Qwen3.5-2B  & \meanstd{62.9}{5.6}  & \meanstd{68.7}{5.2}  & \meanstd{68.6}{5.9}  & \textbf{\meanstd{71.3}{5.4}} \\
Qwen3.5-4B  & \meanstd{84.1}{2.3}  & \meanstd{86.9}{2.6}  & \meanstd{87.7}{1.5}  & \textbf{\meanstd{88.8}{1.5}} \\
Qwen3.5-9B  & \meanstd{86.8}{1.2}  & \meanstd{89.9}{0.5}  & \meanstd{89.6}{1.0}  & \textbf{\meanstd{91.6}{1.0}} \\

\end{tabular}
\end{table}

\subsection{When Constrained Decoding Improve LLM Performance}

The performance gap between JSON and free-form reasoning approaches: InWriting and NL-to-Format, shown in Tables~\ref{tab:NL-to-Format-direct-comparison}, \ref{tab:NL-to-Format-direct-comparison-full}, \ref{tab:gsm8k}, \ref{tab:lastletter}, and \ref{tab:lastletter-extended} is consistent with prior work~\citep{tam-etal-2024-speak, banerjee2025crane, nguyen2026thinking, reddy2026hidden, lee2026format}. However, when the prompt and expected output format are fixed, the combined use of \textbf{Format Constraint (FC)} and \textbf{Validity Constraint (VC)} (+FC+VC) consistently outperforms unconstrained (see Tables~\ref{tab:NL-to-Format-direct-comparison}, \ref{tab:NL-to-Format-direct-comparison-full}, \ref{tab:gsm8k}, \ref{tab:lastletter}, \ref{tab:lastletter-extended}, \ref{tab:lastletter-constrained-gain}, and \ref{tab:lastletter-extended-constrained-gain}). This supports \citet{lee2026format}, who attribute differences across output styles primarily to prompt-level factors.

\begin{figure}[h]
    \centering
    \includegraphics[width=0.7\textwidth]{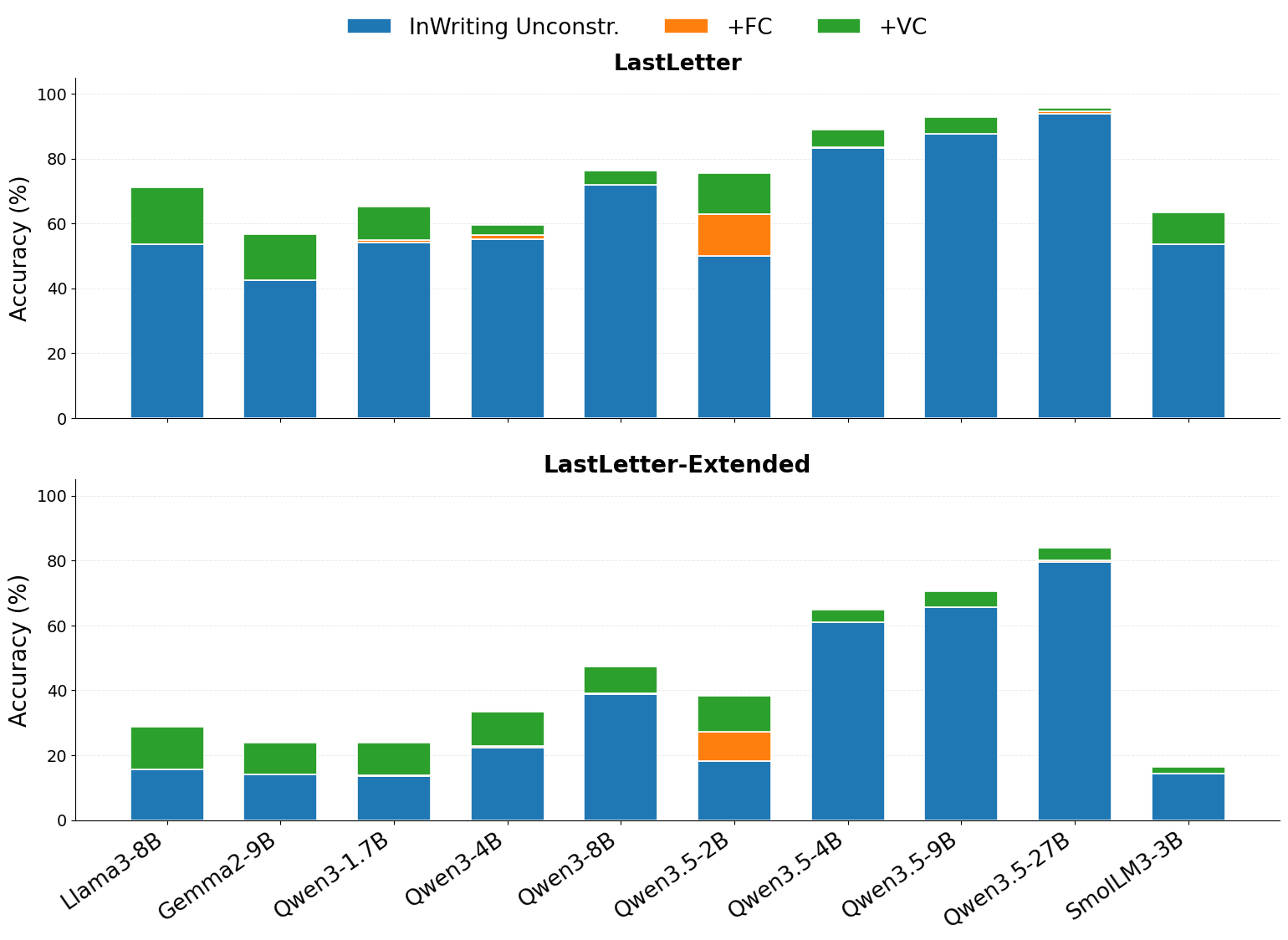}
    \caption{Comparison of the performance gains from Format Constraint (FC) and Validity Constraint (VC) on the LastLetter and LastLetter-Extended tasks.}    \label{fig:gain-fig}
\end{figure}

To further investigate the source of these gains, we consider two levels of constraint: (1) \textbf{Format Constraint (FC)}, which ensures that answers follow the required format and are extractable, and (2) the additional \textbf{Validity Constraint (VC)}, which rules out invalid answers and can turn incorrect answers into correct ones. Results in Fig \ref{fig:gain-fig}, Tables~\ref{tab:lastletter-constrained-gain}, and \ref{tab:lastletter-extended-constrained-gain} show that VC provides more consistent gains than FC (up to 17.4\%). This is because VC provides semantic information about the ground-truth answer, while FC mainly enforces the output format and has less effect when the model already follows the format instruction well. Details of the semantic constraints used in our experiments are provided in Sec.~\ref{sec:prompt-format-constraints}, while details on how Validity Constraints help LLMs turn incorrect answers into correct ones are discussed in Sec.~\ref{sec:vc-sample-level-gain-analysis}, and demonstrate its potential for broader applications through a preliminary experiment on prohibited content filtering in Table~\ref{tab:imdb} and Sec.~\ref{sec:imdb}.

\subsection{Can Reasoning Constraints Benefit LLM Reasoning?}

Beyond post-hoc validity constraints, reasoning constraints (\textbf{RC}) can potentially improve LLM reasoning and accuracy. As shown in Table~\ref{tab:gsm8k-rc} and \ref{tab:aime2026-rc}, enforcing LLMs to begin with ``Let's think step by step'' generally improves performance on GSM8K and AIME2026, consistent with prior work \citep{kojima2022large}. However, the 6.6\% degradation on AIME2026 with Qwen3.5-4B highlights that RC benefits are model- and task-dependent. This suggests that RC is not universally practical: (1) its benefits depend on the task and model; (2) specific constraints may exclude alternative valid reasoning paths; and (3) identifying beneficial traces may require unavailable oracle knowledge.

\subsection{Parsability Analysis}

We analyze the parsability of each decoding strategy in Tables~\ref{tab:gsm8k-parsability} and \ref{tab:lastletter-extended-parsability}, with a sample-level analysis provided in Sec.~\ref{sec:sample-level-failed-parse}. Across all strategies, constrained decoding consistently improves format completion compared with unconstrained decoding. This improvement remains consistent even for larger models, such as Qwen3-32B and Qwen3.5-27B, suggesting that constrained decoding provides reliable control over output formatting regardless of model size.

\subsection{Token Efficiency Analysis}

We report the generated token counts for each decoding method in Table~\ref{tab:token-count}.

Although additional tokens are modest compared with unconstrained decoding, token count does not fully capture computational cost. Constrained decoding requires determining the allowed tokens and applying token-level masking at each step; for JSON, this overhead begins at the start and persists throughout generation, growing with constraint complexity \citep{geng2025generating}. NL-to-Format requires an additional LLM call for parsing, increasing computation and token usage. InWriting adds fewer than 15 tokens in our experiments and applies constrained decoding only to the final output, reducing constraint overhead while preserving free-form reasoning and providing a balance between reasoning flexibility and computational efficiency.

\section{Conclusion}

While the degradation between reasoning in a structured format (JSON) and free reasoning is well documented in prior work and observed in our experiments, we show theoretically and empirically that it arises from the prompt-level constraint rather than constrained decoding itself. In fact, constrained decoding can improve performance when the constraint is appropriately designed.

Beyond guaranteeing the output format, we show that \emph{validity constraints} (VC) provide a near-guaranteed, near-zero-cost gain when the subspace containing valid answers can be characterized from the task structure. By eliminating invalid outputs, VC can improve both syntactic and semantic validity without requiring prior knowledge of the correct answer. This principle naturally extends to applications such as enforcing valid molecular structures in chemistry or filtering prohibited content.

\emph{Reasoning constraints} (RC), however, show mixed results across models and tasks. They can improve performance when the enforced reasoning traces are beneficial, but degrade performance otherwise, highlighting the difficulty of identifying beneficial reasoning traces in practice.

Finally, constrained decoding introduces computational and token-efficiency costs that grow with structural complexity. Reasoning inside JSON requires constraints to be enforced throughout generation, while NL-to-Format requires a second parsing step and nearly doubles token usage. Mode switching through trigger tokens, such as InWriting, offers a lower-overhead alternative that preserves free-form reasoning while constraining only the desired output format. 

This work has limitations that motivate future research across a broader range of domains, tasks, and constraint types, as discussed in Sec.~\ref{sec:limitations-solutions}.

\subsection*{AI use statement}

In this work, we used generative AI tools, primarily the free versions of ChatGPT and Claude, to draft and edit parts of the manuscript for improved readability.

All research questions, ideas, concepts, and experiments were developed and conducted by the authors.

We take full responsibility for the final content of this work, including all text, claims, and artifacts produced with the assistance of generative AI.




\subsection*{Reproducibility statement}








\paragraph{Clarification of Assumptions 1 and 2 in Sec.~\ref{sec:methodology}}

Assumption 1 guarantees non-degradation, while both Assumptions 1 and 2 are required for a performance gain. Both are practical to satisfy. For Assumption 1, we set all fields (e.g., in JSON) to \texttt{str} under the same prompt and expected output format, allowing all possible generations. Thus, if the LLM already follows the predefined structure, constrained decoding has no effect.

Assumption 2 is similarly practical. If the LLM generates an invalid syntax token, such as \texttt{\{}'' or \texttt{"}'', the unconstrained output becomes unextractable, while constrained decoding prevents such tokens and preserves extractability. Importantly, the potential gain increases when more information about the correct answer space is known in advance through the given task, question, or other context.

\paragraph{Clarification on LastLetter-Extended Dataset Construction} Details are provided in Sec~\ref{subsubsec:dataset-construction} and included in the Supplementary Materials.



\bibliography{iclr2027_conference}
\bibliographystyle{iclr2027_conference}

\appendix
\section{Appendix}

\subsection{Limitations and Future Work}\label{sec:limitations-solutions}

Besides the well-known benefit of constrained decoding in guaranteeing the expected output format, our findings highlight two additional benefits: \emph{validity constraints} (VC), which provide a near-guaranteed, near-zero-cost gain by eliminating invalid outputs, and \emph{reasoning constraints} (RC), whose benefits depend on whether the enforced reasoning trajectory is beneficial.

For VC, our main limitation is the narrow experimental scope. While VC can naturally extend to domains such as chemistry, where constraints can enforce valid molecular structures, and content filtering, where constraints can prevent prohibited words or patterns, we evaluate VC primarily on LastLetter and LastLetter-Extended, with a preliminary content-filtering experiment on IMDb. Expanding VC to these domains remains an important direction for future work.

For RC, we show that enforcing beneficial reasoning paths can improve reasoning and answer quality. However, identifying such paths requires prior information, so our experiments demonstrate feasibility using patterns from prior work rather than providing practical guidance for finding them.

Lastly, our experiments use relatively simple JSON formats, consistent with prior work. More complex tasks that require both extensive reasoning and complex output formats remain an important direction for future work.

\subsection{Prompt, Expected Format, and Constraint Variations}
\label{sec:prompt-format-constraints}
We use a prompt template consisting of a task description, format description, few-shot examples, and a question, following prior work \citep{tam-etal-2024-speak,nguyen2026thinking}.

\begin{tcolorbox}
\textbf{Prompt Structure}

\verb|{task_description}|\\
\verb|{format_description}|\\
\verb|{fewshots}|\\
\verb|{question}|
\end{tcolorbox}

We use 3 task-description variations per task, except AIME2026, which uses 1. JSON uses 1 fixed format, while other settings use 3. Each AIME2026 prompt combination is evaluated with 3 seeds.

\begin{itemize}
\item \textbf{GSM8K, LastLetter, LastLetter-Extended -- JSON.}
$3 \times 1 = 3$ prompt combinations.

\item \textbf{GSM8K, LastLetter, LastLetter-Extended -- InWriting, NL-to-Format.}
$3 \times 3 = 9$ prompt combinations.

\item \textbf{AIME2026 -- InWriting.}
$1 \times 3 = 3$ prompt combinations $\times$ 3 seeds $= 9$ evaluation runs.
\end{itemize}

\subsubsection{Prompt Variants}

For \textbf{GSM8K}, \textbf{LastLetter}, and \textbf{LastLetter-Extended}, we use the same task-description variations across all three decoding strategies: \textbf{JSON}, \textbf{InWriting}, and \textbf{NL-to-Format/NL-to-JSON}, following prior work \citep{tam-etal-2024-speak,nguyen2026thinking}, as follows:

\begin{tcolorbox}[colback=green!10, colframe=green!50!black, boxrule=0.5pt, arc=2mm]

\textbf{Task-description Variations for GSM8K}\\[2mm]

\textbf{Variation 1}\\
Follow the instruction to complete the task. 

You are a math tutor who helps students of all levels understand and solve mathematical problems. 

Read the last question carefully and think step by step before answering, the final answer must be only a number.\\[2mm]

\textbf{Variation 2}\\
Follow the instruction to complete the task. 

Read the last question carefully and think step by step before answering, the final answer must be only a number. 

You are a math tutor who helps students of all levels understand and solve mathematical problems.\\[2mm]

\textbf{Variation 3}\\
Follow the instruction to complete the task. 

Mathematical problem-solving task: 

Given: A mathematical question or problem 

Required: A numerical answer only 

Role: You are a math tutor assisting students of all levels 

Process: Think step by step to solve the problem 

Note: Read the question carefully before beginning your analysis.

\end{tcolorbox}

\begin{tcolorbox}[colback=green!10, colframe=green!50!black, boxrule=0.5pt, arc=2mm]

\textbf{Task-description Variations for LastLetter (Both original and Extended) -- Used for all decoding strategies}\\[2mm]

\textbf{Variation 1}\\
Follow the instruction to complete the task. 

You are given a string of words and you need to take the last letter of each word and concatenate them. 

Read the last question carefully and think step by step before answering.\\[2mm]

\textbf{Variation 2}\\
Follow the instruction to complete the task. 

Read carefully for each of the last question and think step by step before answering. 

You are given a string of words and you need to take the last letter of each word and concatenate them.\\[2mm]

\textbf{Variation 3}\\
Follow the instruction to complete the task. 

String manipulation task: 

Given: a sequence of words.

Required: A new string made from the last letter of each word.

Process: Think step by step to solve this challenge.

Note: Ensure you've read the question thoroughly before beginning.

\end{tcolorbox}

For \textbf{AIME2026}, we adopt the task description from \citet{dekoninck2026matharena}, with one modification: we remove the instruction
``Put your final answer within \textbackslash{}boxed\{\{\}\}.''
This format is specific to answer extraction in the original work and is not compatible with all of our decoding strategies.

\begin{tcolorbox}[colback=green!10, colframe=green!50!black, boxrule=0.5pt, arc=2mm]
\textbf{Task Description for AIME2026}\\[2mm]
The answer is an integer between 0 and 999, inclusive.
\end{tcolorbox}

For Format description, we apply the same for all tasks based on the expected output format of each decoding strategies:

For \textbf{JSON} decoding strategy:

\begin{tcolorbox}[colback=blue!5, colframe=blue!50!black, boxrule=0.5pt, arc=2mm]

\textbf{Format Specification for JSON Decoding}\\

Provide your output in the following valid JSON format:

\begin{verbatim}
{
"reason": <think step by step>,
"answer": <answer>
}
\end{verbatim}

\end{tcolorbox}

For \textbf{NL-to-Format} decoding strategy:

\begin{tcolorbox}[colback=blue!5, colframe=blue!50!black, boxrule=0.5pt, arc=2mm]

\textbf{Format Variations for NL-to-Format Decoding Strategies} \\

\textbf{Variation 1}\\
Instruct: Provide your output in the following text format:\\
Answer: \textless reasoning first\textgreater. The final answer is \textless answer\textgreater\\[2mm]

\textbf{Variation 2}\\
Instruct: Provide your output in the following text format:\\
Step by step reasoning: ...\\
Answer: The final answer is ...\\[2mm]

\textbf{Variation 3}\\
Instruct: Provide your output in the following text format:\\
Answer: \textless think step by step\textgreater. The final answer is \textless answer\textgreater

\end{tcolorbox}
For \textbf{InWriting} decoding strategy, we keep the same variations as \textbf{NL-to-Format}, with an additional phrase added to each variation to specify the expected output format for the \textbf{InWriting} decoding strategy.

\begin{tcolorbox}[colback=blue!5, colframe=blue!50!black, boxrule=0.5pt, arc=2mm]

\textbf{Format Variations for InWriting Decoding Strategies}\\

\textbf{Variation 1}\\
Instruct: Provide your output in the following text format:\\
Answer: \textless reasoning first\textgreater. The final answer is \textless answer\textgreater\
\begin{verbatim}
Then, provide the final answer in valid JSON format: 
{"answer": <answer>}
\end{verbatim}\\

\textbf{Variation 2}\\
Instruct: Provide your output in the following text format:\\
Step by step reasoning: ...\\
Answer: The final answer is ...
\begin{verbatim}
Then, provide the final answer in valid JSON format: 
{"answer": <answer>}
\end{verbatim}\\

\textbf{Variation 3}\\
Instruct: Provide your output in the following text format:\\
Answer: \textless think step by step\textgreater. The final answer is \textless answer\textgreater\
\begin{verbatim}
Then, provide the final answer in valid JSON format: 
{"answer": <answer>}
\end{verbatim}

\end{tcolorbox}

\subsubsection{Expected Format and Constraint Variations}

\paragraph{Expected Format.} The expected format differs across strategies. For \textbf{JSON}, the expected format is a JSON object with \texttt{reason} and \texttt{answer} fields. For \textbf{InWriting}, it consists of free-form reasoning followed by a JSON \texttt{answer}, while for \textbf{NL-to-Format}, it consists of free-form reasoning in the first step followed by a JSON \texttt{answer} in the second. We evaluate each strategy in two settings: \emph{unconstrained}, where only the expected format is specified, and \emph{constrained}, where the format is additionally enforced through constrained decoding.

\paragraph{Constraint Variations.}
For \emph{format constraints} (\textbf{FC}), the goal is to enforce only the output format. We therefore set \texttt{answer} to \texttt{str}, allowing any possible answer to be generated. For \emph{validity constraints} (\textbf{VC}), we additionally encode prior knowledge about the potential answer space, based on the task or question, directly into the output schema. For \emph{reasoning constraints} (\textbf{RC}), we enforce beneficial reasoning traces by encoding the desired reasoning pattern into the constraint.

The VC configurations used in our experiments are:
\begin{itemize}
\item \textbf{GSM8K}: The answer is restricted to a non-negative integer.
\item \textbf{AIME2026}: The answer is restricted to an integer between 0 and 999.
\item \textbf{LastLetter-Original}: The answer is restricted to exactly four alphabetic characters.
\item \textbf{LastLetter-Extended}: The answer is restricted to exactly eight alphabetic characters.
\end{itemize}

\paragraph{Output Schema Construction.}
We use Pydantic\footnote{\url{https://docs.pydantic.dev/latest/}} to define and customize the output schemas. The \textbf{JSON} strategy uses two fields, \texttt{reason} and \texttt{answer}, while \textbf{InWriting} and \textbf{NL-to-Format} use only the \texttt{answer} field. For example, the Pydantic schema for the JSON strategy is:
\begin{verbatim}
class Schema(BaseModel):
reason: str
answer: int = Field(ge=0, le=9_999_999_999)
\end{verbatim}

The \texttt{reason} field is fixed as a string, while \texttt{answer} is customized per task. For \emph{validity constraints}, we additionally restrict the answer space by encoding the constraints directly in the field definition, e.g., \verb|int = Field(ge=0, le=9_999_999_999)|. For \emph{format constraints}, we simply set \texttt{answer} to \texttt{str}.

\subsection{Experimental Setup}
\label{sec:experimental-septup}
\subsubsection{Code}
Our code is provided in the Supplementary Material in accordance with the double-blind review policy. The code will be made publicly available upon acceptance.

\subsubsection{Datasets}

All datasets, except for LastLetter-Extended, are open-source and can be directly accessed through Hugging Face Datasets (\href{https://huggingface.co/datasets}{https://huggingface.co/datasets}).

\subsubsection{LastLetter-Extended Dataset Construction}
\label{subsubsec:dataset-construction}

We construct LastLetter-Extended by randomly sampling names from the \texttt{names-dataset} library \href{https://pypi.org/project/names-dataset/}{(https://pypi.org/project/names-dataset/)}, following prior work \citep{wei2022chain, kojima2022large}. We make a minor modification to the sampling procedure to exclude compound names, such as ``Juan Pablo,'' which are treated as a single name in the original dataset without an explicit indication that they consist of multiple names.

\subsubsection{Constrained Decoding Configurations}

We use Litelines\footnote{\url{https://github.com/alonsosilvaallende/litelines/}} for implementing constrained decoding. It provides three key controls: \verb|allow_preamble| for free generation before a trigger, \verb|max_preamble_tokens| to limit unconstrained generation, and \verb|trigger_token_ids| to specify trigger tokens. Following \citep{nguyen2026thinking}, we use EOS and PAD as triggers. These controls enable \textbf{InWriting} to switch from free to constrained generation after a trigger.

An example configuration for \textbf{InWriting} is:
\begin{verbatim}
allow_preamble=True,
max_preamble_tokens=600,
trigger_token_ids=[tokenizer.eos_token_id, tokenizer.pad_token_id]
\end{verbatim}

For \textbf{JSON} and \textbf{NL-to-Format}, we set \verb|allow_preamble=False| to enforce direct generation in the specified format.

\subsubsection{Experimental Configurations}

The configurations of the LLMs used across all experiments are detailed in Table~\ref{tab:experimental-configurations}.
\begin{table}[h!]
\caption{Experimental configurations for the two experimental settings.}
\label{tab:experimental-configurations}
\centering
\small
\begin{tabular}{llll}
\toprule
\textbf{Experiment Type} &
\textbf{Parameter} &
\textbf{Value} &
\textbf{Note} \\
\midrule

\multirow{5}{*}{Standard Setting}
& Temperature & 0 & \\
& Sampling & False & \\
& Max Reasoning Tokens & 600 & \\
& Thinking Mode & Off & \\
& Random Seed & 1 & \\

\midrule

\multirow{5}{*}{\makecell{Extended-Reasoning Setting\\(AIME-2026 only)}}
& Temperature & Default & \\
& Sampling & Default & \\
& Max Reasoning Tokens & 12800 & \\
& Thinking Mode & Off & \\
& Random Seed & 42, 43, 44 & \\

\bottomrule
\end{tabular}
\end{table}

\subsection{Full Experimental Results and Analysis}
\label{sec:full-experimental-results}

\subsubsection{Sample-Level Gain Analysis from Validity Constraints Across Tasks}
\label{sec:vc-sample-level-gain-analysis}

For numerical tasks such as GSM8K, validity constraints primarily improve performance by standardizing answers. For example, \texttt{28.00} is normalized to the required representation, \texttt{28}, allowing the correct answer to be matched successfully.

For string-manipulation tasks: LastLetter and LastLetter-Extended, validity constraints correct formatting artifacts and deviations between the answer and the reasoning trace. We observe several representative cases. 

First, validity constraints can recover characters that are incorrectly omitted from the answer:

\begin{lstlisting}[
    basicstyle=\ttfamily\scriptsize,
    breaklines=true,
    columns=fullflexible
]
"Model output": "Answer: To solve this problem, I will iterate through each word in the given string, extract the last letter of each word, and concatenate them. \n\nThe given string is \"Camilo Becky Eliza Rebecca\". \n\nThe last letters of each word are: o, y, a, a.\n\nThe final answer is \"oya\".\n\n{\"answer\": \"oya\"}",
"Unconstrained parsed_answer": "oya",
"Constrained parsed_answer": "oyaa",
\end{lstlisting}

The reasoning trace identifies the last letters as \texttt{o}, \texttt{y}, \texttt{a}, and \texttt{a}, indicating \texttt{oyaa} as the intended answer. The unconstrained parser omits the final \texttt{a}, while the constrained parser recovers the answer.

A similar effect occurs when the model produces repeated characters in the answer:

\begin{lstlisting}[
    basicstyle=\ttfamily\scriptsize,
    breaklines=true,
    columns=fullflexible
]
"Model output": "Answer: To solve this problem, I will iterate through each word in the given string, extract the last letter of each word, and concatenate them. \n\nThe given string is \"Selina Tasha Jarrod Maddy\". \n\nThe last letters of each word are: a, a, d, y.\n\nThe final answer is \"aaday\".\n\n{\"answer\": \"aaday\"}",
"Unconstrained parsed_answer": "aaday",
"Constrained parsed_answer": "aady",
\end{lstlisting}

Here, the reasoning trace identifies the last letters as \texttt{a}, \texttt{a}, \texttt{d}, and \texttt{y}, yielding \texttt{aady}. The unconstrained parser retains an additional \texttt{a}, while the constrained parser removes it.

Second, validity constraints can remove formatting artifacts such as spaces between characters:

\begin{lstlisting}[
    basicstyle=\ttfamily\scriptsize,
    breaklines=true,
    columns=fullflexible
]
"Model output": "Answer: To solve this problem, I will iterate through each word in the given string, extract the last letter of each word, and concatenate them. \n\nThe given string is \"Yamileth Dane Aron Dee\". \n\nThe last letters of each word are: \n- Yamileth: H\n- Dane: E\n- Aron: N\n- Dee: E\n\nThe final answer is: H E N E\n\n{\"answer\": \"H E N E\"}",
"Unconstrained parsed_answer": "h e n e",
"Constrained parsed_answer": "hene"
\end{lstlisting}

In this example, the model correctly identifies \texttt{H}, \texttt{E}, \texttt{N}, and \texttt{E}, but inserts spaces between them. The unconstrained parser preserves these spaces, producing \texttt{h e n e}, while the constrained parser removes them, producing \texttt{hene}.

While these gains may appear to result from simple post-hoc processing, the token-level behavior reveals a more substantive role for VC. In these examples, spaces and additional characters are not necessarily independent tokens; rather, they are part of tokens such as \texttt{` e'}, \texttt{` n'}, and \texttt{`da'}. Thus, under VC, the model must select tokens that satisfy the validity constraint while remaining consistent with the reasoning trace. From this perspective, VC acts as a post-hoc consistency enforcement rather than merely a processing step, guiding the model toward a valid continuation based on both its reasoning trace and the validity constraints. This perspective extends constrained decoding beyond format enforcement to a broader range of applications, such as enforcing valid molecular structures in chemistry and filtering prohibited content.

\subsubsection{Preliminary Experiment on Prohibited Content Filtering with IMDB}
\label{sec:imdb}

Motivated by findings regarding validity constraints as a post-hoc consistency enforcement, we conduct a preliminary experiment on the IMDb Large Movie Review Dataset \citep{maas-EtAl:2011:ACL-HLT2011}. We investigate whether an LLM can perform sentiment classification when the standard labels, \textit{Positive} and \textit{Negative}, are specified in the prompt but prohibited from generation.

We evaluate three conditions: (1) an \textbf{Unconstrained Baseline} with no decoding restrictions; (2) \textbf{Label-Constrained Decoding}, where the output is restricted to \textit{Positive} or \textit{Negative}; and (3) \textbf{Alternative-Constrained Decoding}, where the output is restricted to a predefined set of sentiment-related alternatives:

\begin{itemize}
\item \textbf{Negative:} Bad, Critical, Unfavorable, Disappointing, Unsatisfactory.
\item \textbf{Positive:} Good, Favorable, Impressive, Satisfying, Enjoyable.
\end{itemize}

The third condition tests whether the model can retain the correct sentiment when constrained to alternative expressions rather than the original class labels.

\textbf{Prompt for IMDb}\
\begin{tcolorbox}[colback=green!10, colframe=green!50!black, boxrule=0.5pt, arc=2mm]

Classify the sentiment of the following text as either Positive or Negative.

Return exactly one word: Positive or Negative.

Do not include explanations, punctuation, or any other text.

\end{tcolorbox}

\begin{table}[h!]
\caption{IMDb sentiment classification under different decoding constraints, comparing the Unconstrained Baseline, Label-Constrained Decoding, and Alternative-Constrained Decoding.}
\label{tab:imdb}
\centering
\small
\begin{tabular}{lccc}

\textbf{Model} &
\makecell{\textbf{Unconstrained}\\\textbf{Baseline}} &
\makecell{\textbf{Label-}\\\textbf{Constrained}} &
\makecell{\textbf{Alternative-}\\\textbf{Constrained}} \\
\midrule
Qwen3-1.7B & 81.6 & 81.6 & 76.0 \\
Qwen3-4B & 92.6 & 92.6 & 83.4 \\
Qwen3.5-2B & 89.6 & 89.6 & 80.8 \\
Qwen3.5-4B & 94.6 & 94.8 & 85.8 \\
\
\end{tabular}
\end{table}

As shown in Table~\ref{tab:imdb}, although Alternative-Constrained Decoding does not match Label-Constrained Decoding, it maintains substantial performance across models. This suggests that LLMs can preserve the intended answer using alternative expressions when the canonical label is prohibited, which is promising for content filtering.

\subsubsection{Sample-Level Analysis of Failed Parses}
\label{sec:sample-level-failed-parse}

While InWriting and NL-to-Format achieve relatively high parse rates without constrained decoding, JSON-format generation suffers substantially in the absence of constraints (see Table~\ref{tab:gsm8k-parsability} and \ref{tab:lastletter-extended-parsability}). Beyond cases where the expected JSON structure is incorrect or missing, we identify two main output patterns in the JSON strategy that lead to parsing failures and need to be discussed further.

\textbf{Case 1: Free-form Generation with JSON at the End}

\begin{lstlisting}[
    basicstyle=\ttfamily\scriptsize,
    breaklines=true,
    columns=fullflexible
]
"Model output": "### Step-by-step Explanation:\n\n1. **Identify the input string**:  \n   The input string is `\"Homoh Aniba Grewar Manying Mordoch Bjarndis Leveron Rahymoff\"`.\n\n2. **Split the string into individual words**:  \n   The string is split into words using spaces as delimiters:  \n   `[\"Homoh\", \"Aniba\", \"Grewar\", \"Manging\", \"Mordoch\", \"Bjarndis\", \"Leveron\", \"Rahymoff\"]`.\n\n3. **Extract the last letter of each word**:  \n   For each word, take the last character:\n   - \"Homoh\" \u2192 \"h\"\n   - \"Aniba\" \u2192 \"a\"\n   - \"Grewar\" \u2192 \"r\"\n   - \"Manging\" \u2192 \"g\"\n   - \"Mordoch\" \u2192 \"h\"\n   - \"Bjarndis\" \u2192 \"s\"\n   - \"Leveron\" \u2192 \"n\"\n   - \"Rahymoff\" \u2192 \"f\"\n\n4. **Concatenate the last letters**:  \n   Combine the extracted letters into a single string:  \n   `\"hagr hsf n f\"`.\n\n### Final Answer:\n\n```json\n{\n  \"reason\": \"The input string is split into individual words, and the last letter of each word is extracted and concatenated.\",\n  \"answer\": \"hagr hsf n f\"\n}\n```"
\end{lstlisting}

\textbf{Case 2: JSON wrapped in a Markdown code fence}

\begin{lstlisting}[
    basicstyle=\ttfamily\scriptsize,
    breaklines=true,
    columns=fullflexible
]
"Model output": "```json\n{\n\"reason\": \"To solve this, I will identify each word in the provided string 'Lalesca Hasiic Elrayan Weixu Alzadgali Samphoas Stierlin Thwmas' and extract its last letter. The words and their last letters are: 'Lalesca' -> 'a', 'Hasiic' -> 'c', 'Elrayan' -> 'n', 'Weixu' -> 'u', 'Alzadgali' -> 'i', 'Samphoas' -> 's', 'Stierlin' -> 'n', 'Thwmas' -> 's'. Finally, I will concatenate these letters in order: 'a' + 'c' + 'n' + 'u' + 'i' + 's' + 'n' + 's', resulting in 'acnuisns'.\",\n\"answer\": \"acnuisns\"\n}\n```\n"
\end{lstlisting}

It may seem harsh at first to classify both cases as failures, as their outputs are still extractable, particularly Case 2, where JSON wrapped in a Markdown code fence is commonly accepted. However, neither case follows the output format specified in the prompt or the schema enforced by constrained decoding. If Case 2 is considered acceptable, the schema must explicitly allow this format; otherwise, constrained decoding would eliminate all paths beginning with \texttt{```json}, contradicting Assumption~1 that constrained decoding does not hurt LLM performance (see Sec.~\ref{sec:methodology}). These cases highlight the need to understand constrained decoding and carefully design schemas that allow all output formats that we consider correct.

\subsubsection{Full Experimental Results}


\begin{table}[h!]
\caption{LastLetter Task: comparison of unconstrained and combined Format and Validity Constraints (FC+VC) for plain-text answers using the NL-to-Format decoding strategy with the same prompt and reasoning traces. \textbf{Bold} values indicate the highest accuracy.}
\label{tab:NL-to-Format-direct-comparison-full}
\centering
\small
\begin{tabular}{llll|ll}
\textbf{$n_{\mathrm{shots}}$} &
\textbf{Model} &
\textbf{NL (Baseline)} &
\makecell{\textbf{JSON}\\\textbf{+FC+VC}} &
\makecell{\textbf{NL-to-Format}\\\textbf{Unconstr. (plain-text)}} &
\textbf{+FC+VC} \\

\midrule

\multirow{9}{*}{0}
& Qwen3-1.7B   & \meanstd{33.6}{21.9} & \meanstd{45.4}{19.1} & \meanstd{60.7}{4.3} & \textbf{\meanstd{66.1}{4.8}} \\
& Qwen3-4B     & \meanstd{32.1}{26.2} & \meanstd{41.9}{21.6} & \meanstd{57.5}{7.5} & \textbf{\meanstd{60.4}{7.1}} \\
& Qwen3-8B     & \meanstd{63.8}{17.4} & \meanstd{78.4}{7.9}  & \meanstd{69.4}{11.4} & \textbf{\meanstd{72.4}{10.6}} \\
& Qwen3.5-2B   & \meanstd{22.4}{29.9} & \meanstd{69.1}{2.2}  & \meanstd{62.9}{5.6} & \textbf{\meanstd{68.7}{5.2}}\\
& Qwen3.5-4B   & \meanstd{81.6}{1.6}  & \meanstd{71.6}{28}   & \meanstd{84.1}{2.3} & \textbf{\meanstd{86.9}{2.6}}\\
& Qwen3.5-9B   & \meanstd{85.9}{1.2}  & \meanstd{77}{27}     & \meanstd{86.8}{1.2} & \textbf{\meanstd{89.9}{0.5}} \\
\midrule

\multirow{9}{*}{1}
& Qwen3-1.7B   & \meanstd{54.1}{4.6}  & \meanstd{35.8}{24.6} & \meanstd{58}{3.1}   & \textbf{\meanstd{62.2}{2.5}}\\
& Qwen3-4B     & \meanstd{60.1}{10.5} & \meanstd{21.1}{26.6} & \meanstd{67.6}{4.7} & \textbf{\meanstd{69.9}{3.9}}\\
& Qwen3-8B     & \meanstd{65.1}{8.4}  & \meanstd{64.4}{20.3} & \meanstd{67.2}{6.7} & \textbf{\meanstd{69.7}{6.6}}\\
& Qwen3.5-2B   & \meanstd{17.4}{22.4} & \meanstd{56.6}{1.7}  & \meanstd{39.6}{19}  & \textbf{\meanstd{42.6}{20.6}}\\
& Qwen3.5-4B   & \meanstd{81.9}{2.3}  & \meanstd{63.2}{35.1} & \meanstd{82.7}{1.9} & \textbf{\meanstd{85.6}{1.9}} \\
& Qwen3.5-9B   & \meanstd{85.5}{1.9}  & \meanstd{81.3}{27.3} & \meanstd{86}{1.5}   & \textbf{\meanstd{88.3}{1.1}} \\
\midrule

\multirow{9}{*}{4}
& Qwen3-1.7B   & \meanstd{55}{3.4}    & \meanstd{39}{19.7}   & \meanstd{56.7}{3.5} & \textbf{\meanstd{61.3}{4}}\\
& Qwen3-4B     & \meanstd{75.5}{2.9}  & \meanstd{27.6}{31}   & \meanstd{76.4}{3.4} & \textbf{\meanstd{78.4}{2.4}}\\
& Qwen3-8B     & \meanstd{72.7}{2.4}  & \meanstd{77.5}{3.4}  & \meanstd{73.9}{2.3} & \textbf{\meanstd{75.9}{2.9}}\\
& Qwen3.5-2B   & \meanstd{58.7}{2.7}  & \meanstd{63.5}{1.1}  & \meanstd{60.1}{3.1} & \textbf{\meanstd{66.5}{2.1}}\\
& Qwen3.5-4B   & \meanstd{82}{1.4}    & \meanstd{63.8}{34.4} & \meanstd{83.4}{1.1} & \textbf{\meanstd{85.8}{0.9}}\\
& Qwen3.5-9B   & \meanstd{87.8}{0.9}  & \meanstd{82.1}{26.2} & \meanstd{88.7}{1.6} & \textbf{\meanstd{90.8}{1.4}} \\

\end{tabular}
\end{table}

\begin{table}[t]
\caption{GSM8K: comparison of unconstrained (Unconstr.) and combined Format and Validity Constraints (FC+VC) across JSON, InWriting, and NL-to-Format decoding strategies. \textbf{Bold} values indicate the highest accuracy.}
\label{tab:gsm8k}
\centering
\small
\begin{tabular}{lll|ll|ll}
\textbf{Model} &
\makecell{\textbf{JSON}\\\textbf{Unconstr.}} &

\textbf{+FC +VC} &
\makecell{\textbf{InWriting}\\\textbf{Unconstr.}} &
\textbf{+FC +VC} &
\makecell{\textbf{NL-to-Format}\\\textbf{Unconstr.}} &

\textbf{+FC +VC} \\
\midrule

Llama3-8B   & \meanstd{42.1}{11.8} & \meanstd{71.8}{1.6} & \meanstd{74.7}{1.3} & \textbf{\meanstd{77.8}{1.2}} & \meanstd{76.4}{1.1} & \meanstd{77.1}{1.2} \\
Gemma2-9B     & \meanstd{0}{0} & \meanstd{86.7}{0.3} & \meanstd{82.9}{1.4} & \textbf{\meanstd{86.7}{0.6}} & \meanstd{83.6}{1.6} & \meanstd{84.4}{1.3} \\
Qwen3-1.7B   & \meanstd{17.6} {24.8} & \meanstd{63.4}{0.5} & \meanstd{68.8}{8.5} & \textbf{\meanstd{75.8}{0.8}} & \meanstd{72.9}{1.1} & \meanstd{74.8}{1.1} \\
Qwen3-4B     & \meanstd{85.3}{1.2} & \meanstd{88.3}{0.2} & \meanstd{85.8}{3.0} & \textbf{\meanstd{89.7}{0.4}} & \meanstd{86.0}{1.1} & \meanstd{87.4}{0.8} \\
Qwen3-8B     & \meanstd{56.7}{22.4} & \meanstd{90.9}{0.4} & \meanstd{89.4}{0.9} & \textbf{\meanstd{91.4}{0.5}} & \meanstd{88.8}{1.6} & \meanstd{90.1}{1.2} \\
Qwen3.5-2B   & \meanstd{61}{1.2} & \meanstd{69.9}{0.9} & \meanstd{47.1}{28.4} & \textbf{\meanstd{74.9}{2.2}} & \meanstd{69.0}{4.6} & \meanstd{72.1}{4.0} \\
Qwen3.5-4B   & \meanstd{88.8}{2.2} & \meanstd{90.6}{0.4} & \meanstd{88.2}{1.5} & \textbf{\meanstd{91.8}{0.4}} & \meanstd{88.7}{1.7} & \meanstd{90.6}{0.9} \\
Qwen3.5-9B   & \meanstd{91.2}{1.5} & \meanstd{91.9}{0.4} & \meanstd{90.4}{1.8} & \textbf{\meanstd{93.9}{0.3}} & \meanstd{91.1}{1.6} & \meanstd{92.6}{1.0} \\
SmolLM3-3B   & \meanstd{76.7}{1.6} & \meanstd{77.1}{1.5} & \meanstd{78.8}{2.5} & \textbf{\meanstd{82.8}{0.6}} & \meanstd{81.8}{1.3} & \meanstd{82.4}{1.3} \\

\end{tabular}
\end{table}

\begin{table}[t]
\caption{LastLetter Task: comparison of unconstrained (Unconstr.) and combined Format and Validity Constraints (FC+VC) across JSON, InWriting, and NL-to-Format decoding strategies. \textbf{Bold} values indicate the highest accuracy. \textit{X} indicates that the corresponding experiment is unavailable.}
\label{tab:lastletter}
\centering
\small
\begin{tabular}{lll|ll|ll}
\textbf{Model} &
\makecell{\textbf{JSON}\\\textbf{Unconstr.}} &
\textbf{+FC+VC} &
\makecell{\textbf{InWriting}\\\textbf{Unconstr.}} &
\textbf{+FC+VC} &
\makecell{\textbf{NL-to-Format}\\\textbf{Unconstr.}} &
\textbf{+FC+VC} \\
\midrule

Llama3-8B
& \meanstd{0}{0}
& \meanstd{2.0}{1.4}
& \meanstd{53.6}{5.4}
& \meanstd{71.0}{3.8}
& \meanstd{68.7}{4.4}
& \meanstd{\textbf{71.3}}{4.0} \\

Gemma2-9B
& \meanstd{0}{0}
& \meanstd{43.8}{2.2}
& \meanstd{42.5}{4.1}
& \meanstd{\textbf{56.6}}{4.4}
& \meanstd{51}{5.8}
& \meanstd{54.1}{5.0} \\

Qwen3-1.7B
& \meanstd{0}{0}
& \meanstd{9.8}{1.1}
& \meanstd{54.2}{4.8}
& \meanstd{65.2}{5.1}
& \meanstd{67.3}{4.1}
& \meanstd{\textbf{68.9}}{4.6} \\

Qwen3-4B
& \meanstd{55.6}{5.7}
& \meanstd{56.7}{5.5}
& \meanstd{55.2}{8.0}
& \meanstd{59.6}{7.3}
& \meanstd{60.3}{6.4}
& \meanstd{\textbf{61.4}}{6.5} \\

Qwen3-8B
& \meanstd{75.6}{2.8}
& \meanstd{79.6}{2.8}
& \meanstd{72.0}{4.9}
& \meanstd{76.4}{5.6}
& \meanstd{72.1}{10.5}
& \meanstd{73.5}{10.6} \\

Qwen3.5-2B
& \meanstd{60.9}{1.7}
& \meanstd{69.1}{2.8}
& \meanstd{50.1}{21.9}
& \meanstd{\textbf{75.5}}{2.7}
& \meanstd{68.6}{5.9}
& \meanstd{71.3}{5.4} \\

Qwen3.5-4B
& \meanstd{83.8}{1.1}
& \meanstd{87.3}{0.9}
& \meanstd{83.3}{1.2}
& \meanstd{\textbf{88.8}}{1.4}
& \meanstd{87.7}{1.5}
& \meanstd{\textbf{88.8}}{1.5} \\

Qwen3.5-9B
& \meanstd{23.1}{16.4}
& \meanstd{88.9}{0.3}
& \meanstd{87.6}{0.9}
& \meanstd{\textbf{92.8}}{0.9}
& \meanstd{89.6}{1.0}
& \meanstd{91.6}{1.0} \\

Qwen3.5-27B
& \meanstd{90.4}{2.2}
& \meanstd{93.3}{0.5}
& \meanstd{93.9}{0.8}
& \meanstd{\textbf{95.7}}{0.3}
& X
& X \\

SmolLM3-3B
& \meanstd{0.2}{0.3}
& \meanstd{0.2}{0.3}
& \meanstd{53.6}{3.5}
& \meanstd{\textbf{63.4}}{5.2}
& \meanstd{59.3}{9.7}
& \meanstd{62.4}{10.0} \\

\end{tabular}
\end{table}

\begin{table}[t]
\caption{LastLetter-Extended Task: comparison of unconstrained (Unconstr.) and combined Format and Validity Constraints (FC+VC) across JSON, InWriting, and NL-to-Format decoding strategies. \textbf{Bold} values indicate the highest accuracy. \textit{X} indicates that the corresponding experiment is unavailable.}
\label{tab:lastletter-extended}
\centering
\small
\begin{tabular}{lll|ll|ll}
\textbf{Model} &
\makecell{\textbf{JSON}\\\textbf{Unconstr.}} &
\textbf{+FC+VC} &
\makecell{\textbf{InWriting}\\\textbf{Unconstr.}} &
\textbf{+FC+VC} &
\makecell{\textbf{NL-to-Format}\\\textbf{Unconstr.}} &
\textbf{+FC+VC} \\
\midrule

Llama3-8B   & \meanstd{0}{0} & \meanstd{2.4}{3.5} & \meanstd{15.5}{3.6} & \meanstd{28.7}{5.8} & \meanstd{28.8}{9.6} & \meanstd{\textbf{31.9}}{10.4} \\

Gemma2-9B  & \meanstd{0}{0} & \meanstd{17.6}{0.3} & \meanstd{14.0}{1.8} & \meanstd{23.9}{4.1} & \meanstd{24.0}{3.1} & \meanstd{\textbf{25.6}}{3.0} \\

Qwen3-1.7B & \meanstd{0}{0} & \meanstd{0}{0} & \meanstd{13.5}{4.8} & \meanstd{23.8}{6.9} & \meanstd{27.0}{5.6} & \meanstd{\textbf{28.4}}{5.3} \\

Qwen3-4B   & \meanstd{15.3}{0.0} & \meanstd{19.1}{0.6} & \meanstd{22.4}{6.4} & \meanstd{\textbf{33.4}}{7.8} & \meanstd{26.1}{7.7} & \meanstd{27.5}{7.3} \\

Qwen3-8B   & \meanstd{34.4}{2.7} & \meanstd{39.6}{3.5} & \meanstd{38.7}{5.4} & \meanstd{\textbf{47.3}}{6.2} & \meanstd{42.1}{7.2} & \meanstd{43.5}{6.9} \\

Qwen3.5-2B & \meanstd{15.1}{4.4} & \meanstd{20.4}{5.4} & \meanstd{18.1}{9.3} & \meanstd{\textbf{38.4}}{8.2} & \meanstd{34.4}{11.3} & \meanstd{36.5}{10.0} \\

Qwen3.5-4B & \meanstd{61.1}{2.6} & \meanstd{63.1}{3.1} & \meanstd{61.0}{2.5} & \meanstd{\textbf{64.8}}{2.1} & \meanstd{58.4}{1.9} & \meanstd{59.6}{2.2} \\

Qwen3.5-9B & \meanstd{41.3}{29.0} & \meanstd{62.7}{5.8} & \meanstd{65.7}{1.4} & \meanstd{\textbf{70.5}}{1.4} & \meanstd{66.7}{2.8} & \meanstd{68.2}{3.3} \\

Qwen3.5-27B & \meanstd{69.3}{15.1} & \meanstd{81.1}{1.4} & \meanstd{79.5}{1.9} & \meanstd{\textbf{83.9}}{1.4} & X & X \\

SmolLM3-3B & \meanstd{0}{0} & \meanstd{0}{0} & \meanstd{14.4}{5.9} & \meanstd{16.5}{6.5} & \meanstd{17.9}{6.9} & \meanstd{\textbf{18.8}}{7.3} \\

\end{tabular}
\end{table}

\begin{table}[t]
\caption{LastLetter Task: decomposing constrained decoding gains relative to unconstrained (Unconstr.) into Format Constraint (FC) and Validity Constraint (VC) across JSON and InWriting decoding strategies.}
\label{tab:lastletter-constrained-gain}
\centering
\small
\begin{tabular}{llll|lll}
\textbf{Model} &
\makecell{\textbf{JSON}\\\textbf{Unconstr.}} &
\textbf{+FC} &
\textbf{+FC+VC} &
\makecell{\textbf{InWriting}\\\textbf{Unconstr.}} &
\textbf{+FC} &
\textbf{+FC+VC} \\
\midrule
Llama3-8B     & \meanstd{0}{0}     & \meanstd{1.8}{1.6} & \meanstd{2.0}{1.4}  & \meanstd{53.6}{5.4} & \meanstd{53.6}{5.4} & \meanstd{71.0}{3.8} \\
Gemma2-9B     & \meanstd{0}{0}     & \meanstd{35.8}{2.7} & \meanstd{43.8}{2.2} & \meanstd{42.5}{4.1} & \meanstd{42.6}{4.0} & \meanstd{56.6}{4.4} \\
Qwen3-1.7B    & \meanstd{0}{0}     & \meanstd{8.4}{1.4} & \meanstd{9.8}{1.1}  & \meanstd{54.2}{4.8} & \meanstd{55.0}{5.4} & \meanstd{65.2}{5.1} \\
Qwen3-4B      & \meanstd{55.6}{5.7} & \meanstd{55.8}{5.6} & \meanstd{56.7}{5.5} & \meanstd{55.2}{8.0} & \meanstd{56.4}{6.9} & \meanstd{59.6}{7.3} \\
Qwen3-8B      & \meanstd{75.6}{2.8} & \meanstd{75.6}{2.8} & \meanstd{79.6}{2.8} & \meanstd{72.0}{4.9} & \meanstd{72.0}{4.9} & \meanstd{76.4}{5.6} \\
Qwen3.5-2B    & \meanstd{60.9}{1.7} & \meanstd{61.6}{2.7} & \meanstd{69.1}{2.8} & \meanstd{50.1}{21.9} & \meanstd{63.0}{2.2} & \meanstd{75.5}{2.7} \\
Qwen3.5-4B    & \meanstd{83.8}{1.1} & \meanstd{83.8}{1.1} & \meanstd{87.3}{0.9} & \meanstd{83.3}{1.2} & \meanstd{83.6}{1.4} & \meanstd{88.8}{1.4} \\
Qwen3.5-9B    & \meanstd{23.1}{16.4} & \meanstd{86.7}{0.5} & \meanstd{88.9}{0.3} & \meanstd{87.6}{0.9} & \meanstd{87.6}{0.9} & \meanstd{92.8}{0.9} \\
Qwen3.5-27B   & \meanstd{90.4}{2.2} & \meanstd{92.7}{0.9} & \meanstd{93.3}{0.5} & \meanstd{93.9}{0.8} & \meanstd{ 94.7}{0.4} & \meanstd{95.7}{0.3} \\
SmolLM3-3B    & \meanstd{0.2}{0.3} & \meanstd{0.2}{0.3} & \meanstd{0.2}{0.3} & \meanstd{53.6}{3.5} & \meanstd{53.6}{3.5} & \meanstd{63.4}{5.2} \\
\end{tabular}
\end{table}

\begin{table}[t]
\caption{LastLetter-Extended Task: decomposing constrained decoding gains relative to unconstrained (Unconstr.) into Format Constraint (FC) and Validity Constraint (VC) across JSON and InWriting decoding strategies.}
\label{tab:lastletter-extended-constrained-gain}
\centering
\small
\begin{tabular}{llll|lll}
\textbf{Model} &
\makecell{\textbf{JSON}\\\textbf{Unconstr.}} &
\textbf{+FC} &
\textbf{+FC+VC} &
\makecell{\textbf{InWriting}\\\textbf{Unconstr.}} &
\textbf{+FC} &
\textbf{+FC+VC} \\
\midrule

Llama3-8B     & \meanstd{0}{0}       & \meanstd{1.8}{2.5} & \meanstd{2.4}{3.5}   & \meanstd{15.5}{3.6} & \meanstd{15.5}{3.6} & \meanstd{28.7}{5.8} \\
Gemma2-9B     & \meanstd{0}{0}       & \meanstd{15.6}{0.6} & \meanstd{17.6}{0.3}  & \meanstd{14.0}{1.8} & \meanstd{14.0}{1.8} & \meanstd{23.9}{4.1} \\
Qwen3-1.7B    & \meanstd{0}{0}       & \meanstd{0}{0} & \meanstd{0}{0}       & \meanstd{13.5}{4.8} & \meanstd{13.8}{4.9} & \meanstd{23.8}{6.9} \\
Qwen3-4B      & \meanstd{15.3}{0.0}  & \meanstd{15.8}{0.3} & \meanstd{19.1}{0.6}  & \meanstd{22.4}{6.4} & \meanstd{22.8}{6.5} & \meanstd{33.4}{7.8} \\
Qwen3-8B      & \meanstd{34.4}{2.7}   & \meanstd{34.4}{2.7} & \meanstd{39.6}{3.5}  & \meanstd{38.7}{5.4} & \meanstd{39.0}{5.4} & \meanstd{47.3}{6.2} \\
Qwen3.5-2B    & \meanstd{15.1}{4.4}   & \meanstd{15.1}{4.4} & \meanstd{20.4}{5.4}  & \meanstd{18.1}{9.3} & \meanstd{27.2}{6.6} & \meanstd{38.4}{8.2} \\
Qwen3.5-4B    & \meanstd{61.1}{2.6}   & \meanstd{61.1}{2.6} & \meanstd{63.1}{3.1}  & \meanstd{61.0}{2.5} & \meanstd{61.0}{2.5} & \meanstd{64.8}{2.1} \\
Qwen3.5-9B    & \meanstd{41.3}{29.0}  & \meanstd{60.9}{5.8} & \meanstd{62.7}{5.8}  & \meanstd{65.7}{1.4} & \meanstd{65.7}{1.4} & \meanstd{70.5}{1.4} \\
Qwen3.5-27B   & \meanstd{69.3}{15.1}  & \meanstd{79.3}{1.1} & \meanstd{81.1}{1.4}  & \meanstd{79.5}{1.9} & \meanstd{80.0}{2.0} & \meanstd{83.9}{1.4} \\
SmolLM3-3B    & \meanstd{0}{0}        & \meanstd{0}{0} & \meanstd{0}{0}       & \meanstd{14.4}{5.9} & \meanstd{14.4}{5.9} & \meanstd{16.5}{6.5} \\

\end{tabular}
\end{table}

\begin{table}[t]
\caption{GSM8K:  Parse sucess rate analysis across JSON, InWriting, and NL-to-Format decoding strategies, comparing unconstrained with Format Constraints (FC).}
\label{tab:gsm8k-parsability}
\centering
\small
\begin{tabular}{lll|ll|ll}
\textbf{Model} &
\makecell{\textbf{JSON}\\\textbf{Unconstr.}} &
\textbf{+FC} &
\makecell{\textbf{InWriting}\\\textbf{Unconstr.}} &
\textbf{+FC} &
\makecell{\textbf{NL-to-Format}\\\textbf{Unconstr.}} &
\textbf{+FC} \\
\midrule

Llama3-8B    & 58.7 & 99.9 & 99.4 & 100 & 99.3 & 100 \\
Gemma2-9B    & 0.2 & 100 & 99.8 & 100 & 99.7 & 100 \\
Qwen3-1.7B   & 28.5 & 97.3 & 90.7 & 100 & 97.9 & 100 \\
Qwen3-4B     & 96.7 & 99.8 & 97.3 & 100 & 98.2 & 100 \\
Qwen3-8B     & 62.6 & 99.8 & 99.7 & 100 & 98.6 & 100 \\
Qwen3.5-2B   & 82.2 & 88.4 & 63.1 & 99.6 & 95.9 & 100 \\
Qwen3.5-4B   & 98.6 & 98.6 & 95.6 & 99.8 & 97.9 & 100 \\
Qwen3.5-9B   & 98.2 & 98.2 & 96.3 & 99.9 & 98.2 & 100 \\
SmolLM3-3B   & 98.4 & 99.5 & 99.2 & 100 & 99.4 & 100 \\

\end{tabular}
\end{table}

\begin{table}[t]
\caption{LastLetter-Extended:  Parse sucess rate analysis across JSON, InWriting, and NL-to-Format decoding strategies, comparing unconstrained with Format Constraints (FC). \textit{X} indicates that the corresponding experiment is unavailable.}
\label{tab:lastletter-extended-parsability}
\centering
\small
\begin{tabular}{lll|ll|ll}
\textbf{Model} &
\makecell{\textbf{JSON}\\\textbf{Unconstr.}} &
\textbf{+FC} &
\makecell{\textbf{InWriting}\\\textbf{Unconstr.}} &
\textbf{+FC} &
\makecell{\textbf{NL-to-Format}\\\textbf{Unconstr.}} &
\textbf{+FC} \\
\midrule

Llama3-8B    & 0 & 100 & 99.8 & 100 & 99.6 & 100  \\
Gemma2-9B    & 0 & 100 & 100 & 100 & 100 & 100 \\
Qwen3-1.7B   & 0.7 & 96.2 & 97.4 & 100 & 99.6 & 100 \\
Qwen3-4B     & 100 & 100 & 99.7 & 100 & 100 & 100 \\
Qwen3-8B     & 100 & 100 & 99.5 & 100 & 99.6 & 100 \\
Qwen3.5-2B   & 99.6 & 99.6 & 67.9 & 99.6 & 99.6 & 100 \\
Qwen3.5-4B   & 93.8 & 94.0 & 92.1 & 100 & 99.9 & 100 \\
Qwen3.5-9B   & 60.4 & 99.3 & 93.1 & 99.9 & 99.8 & 100 \\
Qwen3.5-27B  & 82.4 & 93.8 & 88.7 & 99.9 & X & X \\
SmolLM3-3B   & 97.6 & 99.8 & 99.5 & 99.9 & 99.7 & 100 \\

\end{tabular}
\end{table}

\begin{table}[t]
\caption{Comparison of generated output token counts for JSON, InWriting, and NL-to-Format decoding strategies, under unconstrained and combined Format and Validity Constraints (FC+VC).}
\label{tab:token-count}
\centering
\small
\begin{tabular}{ll|ll|ll|ll}
\textbf{Tasks} &
\textbf{Model} &
\makecell{\textbf{JSON}\\\textbf{Unconstr.}} &
\makecell{\textbf{+FC}\\\textbf{+VC}} &
\makecell{\textbf{InWriting}\\\textbf{Unconstr.}} &
\makecell{\textbf{+FC}\\\textbf{+VC}} &
\makecell{\textbf{NL-to-Format}\\\textbf{Unconstr.}} &
\makecell{\textbf{+FC}\\\textbf{+VC}} \\
\midrule

\multirow{6}{*}{GSM8K}
& Qwen3-1.7B  & 221.1 & 129.2 & 189.5 & 201.1 & 190.4 & 190.2 \\
& Qwen3-4B    & 135.8 & 135.0 & 164.6 & 172.8 & 161.4 & 161.3 \\
& Qwen3-8B    & 223.0 & 159.0 & 176.4 & 188.2 & 174.4 & 174.3 \\
& Qwen3.5-2B  & 242.2 & 242.8 & 278.4 & 286.9 & 304.2 & 302.2 \\
& Qwen3.5-4B  & 164.2 & 163.5 & 272.8 & 280.6 & 269.8 & 269.7 \\
& Qwen3.5-9B  & 184.7 & 183.7 & 277.5 & 285.2 & 259.5 & 259.4 \\
\midrule

\multirow{6}{*}{LastLetter}
& Qwen3-1.7B  & 192.0 & 73.4 & 145.5 & 157.1 & 151.7 & 151.8 \\
& Qwen3-4B    & 101.4 & 101.5 & 112.6  & 119.8 & 108.0 & 108.1 \\
& Qwen3-8B    & 109.4 & 109.4 & 127.4 & 137.7 & 129.1 & 129.2 \\
& Qwen3.5-2B  & 152.8 & 152.5 & 181.3 & 187.8 & 155.0 & 154.5 \\
& Qwen3.5-4B  & 152.8 & 144.7 & 189.4 & 195.7 & 192.0 & 192.0 \\
& Qwen3.5-9B  & 135.1 & 132.0 & 178.3 & 184.1 & 167.0 & 167.0 \\

\end{tabular}
\end{table}




\begin{table}[t]
\caption{GSM8K performance under the InWriting decoding strategy: comparison of Format and Validity Constraints (FC+VC) with FC+VC combined with Reasoning Constraints (FC+VC+RC). \textbf{Bold} values indicate the highest accuracy in each row.}
\label{tab:gsm8k-rc}
\centering
\small
\begin{tabular}{lll}
\textbf{Model} &
\textbf{+FC+VC} &
\makecell{\textbf{+FC+VC}\\\textbf{+RC}} \\
\midrule

Qwen3-1.7B   & \meanstd{75.8}{0.8} & \meanstd{\textbf{77.7}}{0.9} \\
Qwen3-4B     & \meanstd{\textbf{89.7}}{0.4} & \meanstd{89.6}{0.3} \\
Qwen3-8B     & \meanstd{91.4}{0.5} & \meanstd{\textbf{91.9}}{0.6} \\
Qwen3.5-2B   & \meanstd{74.9}{2.2} & \meanstd{\textbf{77.4}}{0.5} \\
Qwen3.5-4B   & \meanstd{\textbf{91.8}}{0.4} & \meanstd{91.5}{0.5} \\
Qwen3.5-9B   & \meanstd{\textbf{93.9}}{0.3} & \meanstd{\textbf{93.9}}{0.3} \\
SmolLM3-3B   & \meanstd{82.8}{0.6} & \meanstd{\textbf{83.7}}{0.6} \\

\end{tabular}
\end{table}




\begin{table}[t]
\caption{AIME2026 performance under the InWriting decoding strategy: comparison of Format and Validity Constraints (FC+VC) with FC+VC combined with Reasoning Constraints (FC+VC+RC). \textbf{Bold} values indicate the highest accuracy in each row.}
\label{tab:aime2026-rc}
\centering
\small
\begin{tabular}{lll}
\textbf{Model} &
\textbf{+FC+VC} &
\makecell{\textbf{+FC+VC}\\\textbf{+RC}} \\
\midrule

Qwen3-4B    & \meanstd{14.8}{4.5} & \meanstd{\textbf{16.7}}{2.7} \\
Qwen3-8B    & \meanstd{12.2}{2.2} & \meanstd{\textbf{13.3}}{3.1} \\
Qwen3.5-4B  & \meanstd{\textbf{62.2}}{3.1} & \meanstd{55.6}{5.7} \\
Qwen3.5-27B & \meanstd{81.9}{3.9} & \meanstd{\textbf{84.3}}{5.6} \\

\end{tabular}
\end{table}

\end{document}